# Lung Nodule Image Synthesis Driven by Two-Stage Generative Adversarial Networks

Lu Cao, Xiquan He, Junying Zeng, Chaoyun Mai* and Min Luo

Wuyi University, Jiangmen 529020, China

**Abstract.** The limited sample size and insufficient diversity of lung nodule CT datasets severely restrict the performance and generalization ability of detection models. Existing methods generate images with insufficient diversity and controllability, suffering from issues such as monotonous texture features and distorted anatomical structures. Therefore, we propose a two-stage generative adversarial network (TSGAN) to enhance the diversity and spatial controllability of synthetic data by decoupling the morphological structure and texture features of lung nodules. In the first stage, StyleGAN is used to generate semantic segmentation mask images, encoding lung nodules and tissue backgrounds to control the anatomical structure of lung nodule images; The second stage uses the DL-Pix2Pix model to translate the mask map into CT images, employing local importance attention to capture local features, while utilizing dynamic weight multi-head window attention to enhance the modeling capability of lung nodule texture and background. Compared to the original dataset, the accuracy improved by 4.6% and mAP by 4% on the LUNA16 dataset. Experimental results demonstrate that TSGAN can enhance the quality of synthetic images and the performance of detection models.

## 1  Introduction

Lung cancer is one of the most common and deadly forms of cancer worldwide[1], with approximately 350 people dying from it every day globally[2]. The most effective treatment for lung cancer occurs in the early stages, where the 5-year relative survival rate exceeds 60%, compared to less than 5% for advanced lung cancer. Therefore, early detection of lung cancer, minimizing unnecessary tests, and alleviating patient anxiety are of critical importance.

Lung nodule lesions range in size from 3mm to 30mm. Due to the varying sizes, shapes, and densities of lung nodules, as well as their potential location within complex lung structures, they exhibit diverse appearances on imaging, making accurate identification challenging[3,28]. Currently, CT scans are commonly used to scan the thoracic region of patients, and doctors analyze the characteristics of lung nodules in CT images to assess the likelihood of lung cancer in screening subjects. As the number of CT screenings increases, physicians face significant pressure from the volume of image analysis. To assist doctors in more efficiently screening lung nodule images, computer-aided diagnosis (CAD) systems based on deep learning have been widely studied[4-5]. Typically, CAD systems require a sufficient amount of annotated data; however, the current lack of datasets leads to poor model performance. This situation is primarily caused by the following challenges: first, due to privacy protection concerns, there are obstacles to data sharing between hospitals and research institutions; second, differences in CT device parameters across multiple centers result in inconsistent data formats and resolutions, increasing processing difficulty; third, high-quality CT data relies on professional medical staff for annotation, but this work is time-consuming and labor-intensive. Traditional data augmentation methods[6] (such as random rotation, scaling, and mirror flipping) expand the training set through geometric transformations, which can increase the diversity of the dataset to some extent but do not address the fundamental issue of insufficient data.

Generative Adversarial Networks (GANs)[7] have been introduced into the field of medical image synthesis, offering new insights for lung nodule data augmentation. For example, Nishio et al.[8] guided 3D GANs using nodule size attributes to generate CT images of lung nodules of specified sizes, effectively enhancing the diversity of generated images. Han et al. [9] proposed a 3D multi-condition GAN to generate diverse nodules in 3D lung CT images, aiming to enhance the performance of object detection models. However, existing methods still face two main issues: (1) limited diversity in nodule locations and background settings; and (2) low controllability in morphological features.

To address these challenges, this paper proposes a two-stage generative adversarial network (TSGAN) that uses mask-guided control to generate lung nodule images. In the first stage of this method, a semantic segmentation mask image is generated to provide structural prior information. In the second stage, the DL-Pix2Pix model generates lung nodule images under the guidance of the mask image. An attention module is added to the generator to enable the model to focus on the features of lung nodule shapes and backgrounds. This two-stage model improves image diversity and controllability by decoupling the morphology and texture of lung nodule images.

## 2    Related Work

### 2.1    Generative adversarial networks

GAN generators and discriminators are both based on deep neural networks. During training, the generator and discriminator compete with each other, continuously improving their own strategies to form a Nash equilibrium. This gives GANs a significant advantage in generating high-quality images. Early GAN models were limited by issues such as training instability and insufficient generation quality. Researchers have continuously broken through technical bottlenecks through architectural innovation and theoretical improvements. The introduction of convolutional neural networks enables the generator to capture the spatial hierarchical features of images, while the proposal of the Wasserstein distance significantly alleviates the phenomenon of mode collapse[10]. Notably, StyleGAN[11] achieved a breakthrough by decoupling style attributes in the latent space and introducing a layer-wise noise injection mechanism, enabling precise control over fine-grained features (such as texture and shape) in generated images. These advancements have endowed GANs with precise control over images, laying the foundation for image generation.

Recently, the field of medical image processing has developed rapidly with the advancement of deep learning technology. This progress is mainly reflected in the different applications of GAN models, including image super-resolution, data augmentation, and modality conversion[12-14]. Chuquicusma et al.[15] used DCGAN to generate realistic lung nodule images and verified the beneficial effects of synthetic data on diagnostic model training through experiments. However, early unsupervised GAN-based methods struggled to control lesion shape and position. To enhance controllability, researchers gradually introduced different prior conditions (such as class labels, attributes, or images) to constrain the generation process. For example, Liu et al.[16] proposed a method called A-GAN, which guides the generative adversarial network using statistical models, achieving efficient and controllable medical image data augmentation; Liang et al.[17] proposed a generative adversarial network (spGAN) combining sketch guidance and incremental generation strategies to synthesize high-resolution, editable, and realistic ultrasound images, achieving realistic synthesis of high-resolution B-mode ultrasound images.

This study proposes TSGAN, which uses lung nodule mask images to guide the generation of lung nodule images, thereby achieving more accurate anatomical constraints.

**2.2 Image Translation**

The task of generating real images from semantic information masks is commonly referred to as image translation, such as generating real street view images from segmented city street view maps. Image translation has garnered significant attention in recent years, as this method produces images with relatively controllable layouts and diversity, and can help train more robust models. The key to this task lies in using semantic segmentation masks as conditions to generate high-quality images that match them. Models based on GANs for this task include Pix2Pix[18], CycleGAN[19], and SPADE[20]. Pix2Pix, the first to emerge, is a framework based on conditional generative adversarial networks. The core idea of this model is to use labeled maps or black-and-white images as input and output realistic images. Later, CycleGAN was proposed by researchers, introducing a cyclic consistency loss to successfully achieve unsupervised image translation. Pix2PixHD[21] further optimized the network architecture and loss function, significantly improving the resolution and detail of generated images, producing clearer and more refined results compared to the original Pix2Pix model. Subsequently, to further enhance the detail and realism of generated images, the SPADE model was proposed, which introduces spatial adaptive normalization layers in each layer of the generator, making the generated images more consistent with the semantic information of the input in terms of local details.

In the field of medical imaging, many studies have utilized image translation techniques to address data scarcity and modality missing issues. For example, Wang et al.[22] proposed an unsupervised zero-shot learning method for cross-modal medical image translation, demonstrating higher realism and structural consistency in medical image translation tasks. Chen et al.[23] proposed a multi-domain medical image translation model MI-GAN based on key transfer branches, which can convert from normal disease-free images to disease target domains, significantly improving the performance of lung disease classification models. Qiu et al.[24] introduced a model called Siamese-Diffusion, which consists of a mask diffusion branch and an image diffusion branch. Their research shows that the image diffusion branch can effectively constrain the mask diffusion branch, improving the fidelity of the images synthesized by the mask diffusion branch while ensuring diversity and scalability.

We added an attention module to the pix2pix generator so that the model focuses on both overall and local details.

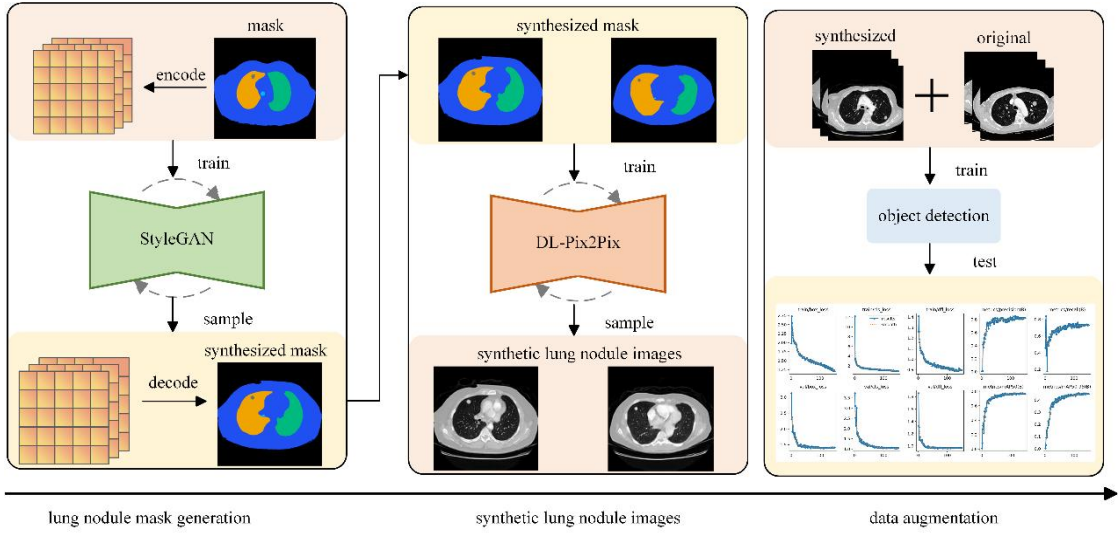

Fig. 1. Overall framework for lung nodule image generation

## 3 TSGAN network

The method proposed in this paper divides the synthesis of lung nodules into four steps: lung nodule mask image generation and lung nodule image generation. Specifically, first, StyleGAN is used to generate lung nodule mask images with diverse morphologies to construct an initial candidate set covering nodule features of different locations and shapes. Then, guided by the lung nodule mask images, the model generates lung nodule images.

### 3.1 Lung mask image generation

To ensure that the distribution of generated lung nodules is similar to that of real lung nodules, we considered both the shape and position of the nodules when generating the lung nodule image mask map. The lung nodule mask map uses different codes to distinguish different regions, where 0, 1, 2, 3, 4, and 5 represent the background, body, left lung, right lung, trachea, and lung nodule region, respectively. In this paper, we use StyleGAN to train the proposed lung nodule image mask map. The trained generator G samples a 512-dimensional latent variable z from a Gaussian distribution N(0, I), then maps z to an embedding space w composed of multiple stages. After passing through a series of convolutional layers, batch normalization, and LeakyReLU activation functions, the resolution of the feature maps is gradually expanded, ultimately generating a 512×512 lung nodule mask image. The discriminator loss $\mathcal{L}_D$ and generator loss $\mathcal{L}_G$ of StyleGAN are expressed as follows:

$$\mathcal{L}_D = \mathbb{E}_{z \sim p(z)}[D(G(z))] - \mathbb{E}_{x \sim p_{data}}[D(x)] + \lambda_{gp} \cdot \mathbb{E}_{\hat{x} \sim p_{interp}}\left[(\|\nabla_x D(\hat{x})\|_2 - 1)^2\right] + \lambda_{drift} \cdot \mathbb{E}\left[D(\cdot)^2\right], \qquad (1)$$

$$\mathcal{L}_G = -\mathbb{E}_{z \sim p(z)}[D(G(z))], \qquad (2)$$

In this equation, $G(z)$ represents the image generated by the generator from the latent variable $z$, $D(\cdot)$ represents the output of the discriminator for all inputs, and $\hat{x}$ is the linear interpolation of the real image and the generated image. $\lambda_{gp}$ and $\lambda_{drift}$ represent the gradient penalty weight and drift regularization weight, respectively.

### 3.2 Lung nodule image generation

To enhance the performance of downstream tasks, this paper uses DL-Pix2Pix to generate lung nodule images. The generator structure is shown in Figure 2. DL-Pix2Pix uses Pix2Pix as the baseline model and incorporates two attention mechanisms into the UNet architecture: Local Importance-based Attention (LIA)[25] and Dynamically Weighted Multi-Head (DWMH) attention. Specifically, LIA attention is incorporated into the skip connections of UNet to enhance the model's focus on local key features. Additionally, DWMH attention is integrated into the bottleneck layer to further capture global contextual information.

**Local Importance-based Attention Mechanism (LIA).** The LIA attention module is a lightweight attention mechanism designed to enhance local features. It aims to address the limitations of traditional attention methods in terms of computational efficiency and multi-scale modeling through dynamic soft pooling and dual-path gating. The input features are modulated by both spatial importance heatmaps and channel gating factors before being output. The core formula is as follows:

$$\text{Output} = \mathbf{X} \odot \mathbf{W} \odot \mathbf{g}, \tag{3}$$

Where $\mathbf{X} \in \mathbb{R}^{C \times H \times W}$ is the input feature, $\mathbf{W} \in [0,1]^{H \times W}$ is the spatial importance heatmap, $\mathbf{g} \in [0,1]^{C}$ is the channel dimension gating factor, and $\odot$ denotes element-wise multiplication.

LIA consists of a dynamic soft pooling layer, a heatmap generation branch, and a gating path. The dynamic soft pooling layer (SoftPooling2D) uses an exponential weighting strategy to adaptively aggregate features, which is mathematically expressed as:

$$\text{SoftPool}(\mathbf{X}) = \frac{\text{AvgPool}(\mathbf{X} \cdot e^{\mathbf{X}})}{\text{AvgPool}(e^{\mathbf{X}})}, \tag{4}$$

The heatmap generation branch extracts multi-scale importance weights through cascaded convolution and pooling operations: the input features are first compressed through a 1×1 convolution layer, then a coarse-grained heatmap is generated through a SoftPooling2D layer with a kernel size of 7 and a stride of 3; the features are then downsampled through a 3×3 convolution layer with a stride of 2, and the channel dimension is restored through another 3×3 convolution layer, finally generating the heatmap $\mathbf{W}$ through a Sigmoid activation function. This design balances receptive fields and computational complexity through hierarchical downsampling.

The gate path performs Sigmoid activation on the first channel of input features to generate channel gate factors $\mathbf{g}$, and upscales the low-resolution heat map $\mathbf{W}$ generated by the branch output to the input size through bilinear interpolation. Finally, the original features $\mathbf{W}$ and $\mathbf{g}$ are multiplied element by element to form fine-grained enhanced features.

**Dynamic Weighted Multi-Head Attention Mechanism (DWMH).** DWMH achieves collaborative modeling of local features and global dependencies by integrating convolution operations with multi-head attention mechanisms. As shown in Figure 3, this module consists of three core components: 1) a convolution-based feature projection layer; 2) a windowed multi-head attention calculation layer; and 3) a dynamic head weight parameterization mechanism.

Given the input feature map $\mathbf{X} \in \mathbb{R}^{C \times H \times W}$, first generate query, key, and value projections through three independent convolution layers:

$$\mathbf{Q}, \mathbf{K}, \mathbf{V} = \text{Conv}_{1 \times 1}(\mathbf{X}), \tag{5}$$

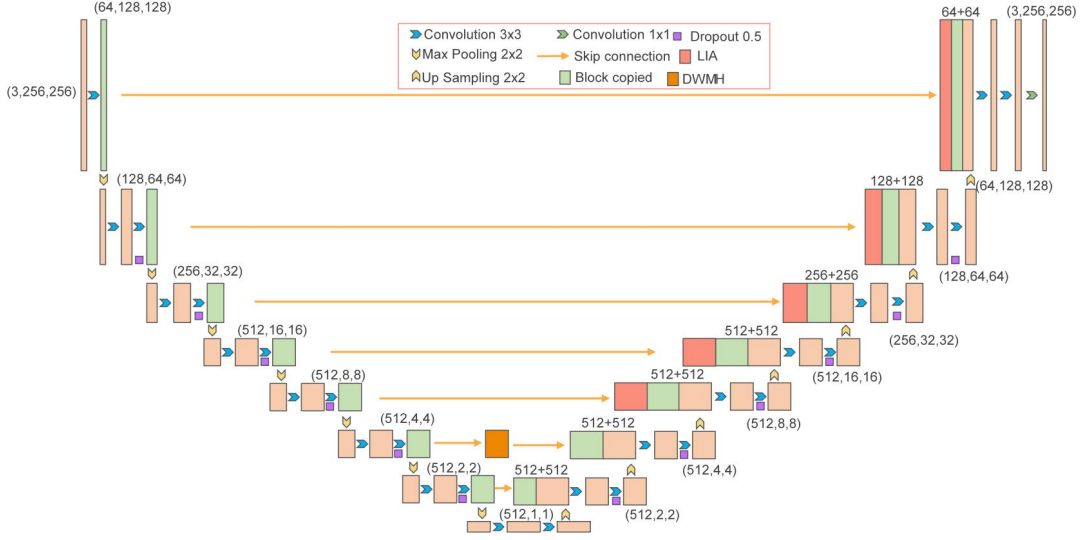

**Fig. 2.** DL-Pix2Pix generator architecture

Use a sliding window strategy to divide the spatial dimension into $N_w = \left\lceil \frac{H}{s} \right\rceil \times \left\lceil \frac{W}{s} \right\rceil$ local windows (where $s$ is the preset window size). For the feature block $\mathbf{X}_i \in \mathbb{R}^{C \times s \times s}$ of the $i$ window, calculate its attention matrix:

$$\text{Attention}_i = \text{Softmax}\left(\frac{\mathbf{Q}_i \mathbf{K}_i^\top}{\sqrt{d_k}}\right) \circ \mathbf{W}_h, \tag{6}$$

Where $d_k = C/n_h$ is the header dimension, $\mathbf{W}_h \in \mathbb{R}^{n_h \times 1 \times 1}$ is the learnable header weight parameter, and $\circ$ represents scalar broadcast multiplication, i.e., multiplying the output of the $h$ attention header by the learnable weight $\mathbf{W}_h^{(h)} \in \mathbb{R}$ and expanding it to the full dimension through the broadcast mechanism. The final output is calculated by dynamically aggregating the results of each attention header:

$$\text{Output}_i = \gamma \cdot (\text{Attention}_i \mathbf{V}_i) + \mathbf{X}_i, \tag{7}$$

In the formula, $\gamma$ is the learnable scaling factor, initialized to 0 to stabilize the early training stage.

This module reduces computational complexity through local window partitioning, improving computational efficiency while maintaining spatial structure perception capabilities.

**Generator Loss Function.** The total loss function of the DL-Pix2Pix generator consists of three parts: adversarial loss, L1 pixel-level reconstruction loss, and perceptual loss. Its mathematical expression is as follows:

$$\mathcal{L}_{\text{Total}} = \mathcal{L}_{\text{GAN}} + \lambda_{\text{L1}} \cdot \mathcal{L}_{\text{L1}} + \lambda_{\text{Perceptual}} \cdot \mathcal{L}_{\text{Perceptual}}, \tag{8}$$

To preserve pixel-level consistency between the generated image and the target image, an L1 norm constraint is introduced. This loss forces the generator to recover high-frequency details, avoiding image blurring issues caused by adversarial training:

$$\mathcal{L}_{\text{L1}} = \mathbb{E}_{x,y}\left[\|y - G(x)\|_1\right], \tag{9}$$

To enhance the semantic reasonableness and structural similarity of the generated images, a pre-trained VGG-19 network is used to extract feature maps, and the distance in the feature space is calculated:

$$\mathcal{L}_{\text{Perceptual}} = \mathbb{E}_{x,y}\left[\sum_{l} \| \phi_l(y) - \phi_l(G(x)) \|_1 \right], \tag{10}$$

Among them, $\phi_l(\cdot)$ represents the feature map of layer $l$ in the VGG-19 network..

## 4 Experiments and analysis

### 4.1 Data preprocessing

This paper uses LUNA16 as the dataset, which contains 888 CT images of lung nodules and includes expert annotations. The size of the lung nodules ranges from 3 to 30 mm. The preprocessing steps mainly involve image normalization and lung parenchyma extraction. Finally, 1,186 lung CT slices were retained. The lung parenchyma extraction results are shown in Figure 3.

This paper uses the same preprocessing method as the original data to ensure data standardization. First, lung parenchyma is extracted. Second, the generated data is used to calculate the corresponding labels in the semantic segmentation map, which contains information on the location of lung nodules. After these two steps, the data is converted to COCO format and input into the detection model.

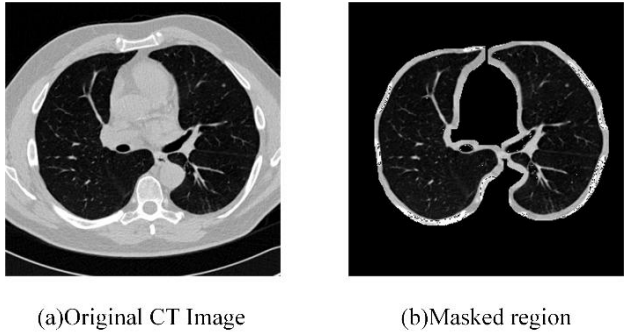

(a)Original CT Image          (b)Masked region

**Fig. 3.** Pulmonary nodule CT image preprocessing

### 4.2 Experimental Platform and Parameter Settings

During the training phase, the input image size was 512×512, and the training set and test set were divided in a 4:1 ratio. Python and the TensorFlow deep learning framework were used to train StyleGAN, while the PyTorch framework was used to train the DL-Pix2Pix model and the lung nodule detection model. The experimental platform is as follows: CPU is Intel i5-12400, memory is 32GB, GPU is NVIDIA GeForce RTX 4070 Super, and CUDA is 12.4.

The specific settings for each part of the experiment are as follows:

(1) StyleGAN employs a progressive growth strategy, gradually increasing the resolution from the initial 4×4 to the target resolution of 512×512. Both the generator G and discriminator D use the Adam optimizer with a learning rate of 0.002, *β1 = 0*, and *β2 = 0.99*. The number of training steps for each resolution is 10,000. In Equation 1, $\lambda_{gp}$ and $\lambda_{drift}$ are set to 10 and 0.001, respectively, in this experiment.

(2) The batch size for training the DL-pix2pix model is 1, using the Adam optimizer with *β1 = 0.5* and a learning rate of 0.0002. The weight for L1 regularization in the loss function is set to 100, with a total of 200 training epochs and a linear learning rate decay strategy starting at the 100th epoch. In Equation 8, $\lambda_{L1} = 200$ and $\lambda_{Perceptual} = 10$.

(3) To validate the impact of synthetic data on the detection model, YOLOv5 and YOLOv8 were selected for lung nodule detection experiments. The batch size was set to 6, and a 5-fold cross-validation strategy was adopted.

## 4.3 Evaluation Criteria

This paper constructs a comprehensive evaluation system from two dimensions: generative quality assessment and detection performance verification. The specific indicator definitions and selection criteria are as follows:

**Generative Quality Assessment Indicators**

(1) Fréchet Inception Distance (FID)

The visual authenticity of the generated image is assessed by calculating the distribution distance between the generated image and the real image in the Inception-v3 feature space. A lower FID value indicates higher generative quality. The calculation formula is:

$$\text{FID} = \| \mu_r - \mu_g \|^2 + \text{Tr}(\Sigma_r + \Sigma_g - 2(\Sigma_r \Sigma_g)^{1/2}), \qquad (11)$$

Where $\mu_r, \Sigma_r \sim \mathcal{N}(\mu_r, \Sigma_r)$ and $\mu_g, \Sigma_g \sim \mathcal{N}(\mu_g, \Sigma_g)$ represent the mean and covariance matrix of the features of the real image and the generated image, respectively.

(2) Peak Signal-to-Noise Ratio (PSNR)

Measures the pixel-level similarity between generated images and real images, defined as:

$$\text{PSNR} = 10 \cdot \log_{10}\left(\frac{\text{MAX}_I^2}{\text{MSE}}\right), \qquad (12)$$

where is the maximum pixel value (4095 for CT images), and MSE is the mean squared error.

(3) Structural Similarity Index (SSIM)

Evaluates image similarity from three dimensions: brightness, contrast, and structure. The SSIM of windows x and y is calculated as:

$$\text{SSIM}(x, y) = \frac{(2\mu_x\mu_y + C_1)(2\sigma_{xy} + C_2)}{(\mu_x^2 + \mu_y^2 + C_1)(\sigma_x^2 + \sigma_y^2 + C_2)}, \qquad (13)$$

In the formula, $\mu$ is the mean value, $\sigma$ is the standard deviation, $C_1 = (0.01L)^2, C_2 = (0.03L)^2$, and $L$ is taken as 255 in this experiment.

**Detection Performance Evaluation Metrics.** Precision, recall, and mean average precision (mAP) are used as evaluation metrics for detection. The expressions for accuracy, recall, and mAP are:

$$\text{Precision} = \frac{TP}{TP + FP}, \qquad (14)$$

$$\text{Recall} = \frac{TP}{TP + FN}, \qquad (15)$$

$$\text{mAP} = \frac{1}{10} \sum_{k=1}^{10} AP_{0.5 + 0.05k}, \qquad (16)$$

where the IoU threshold range is [0.5, 0.95] with a step size of 0.05.

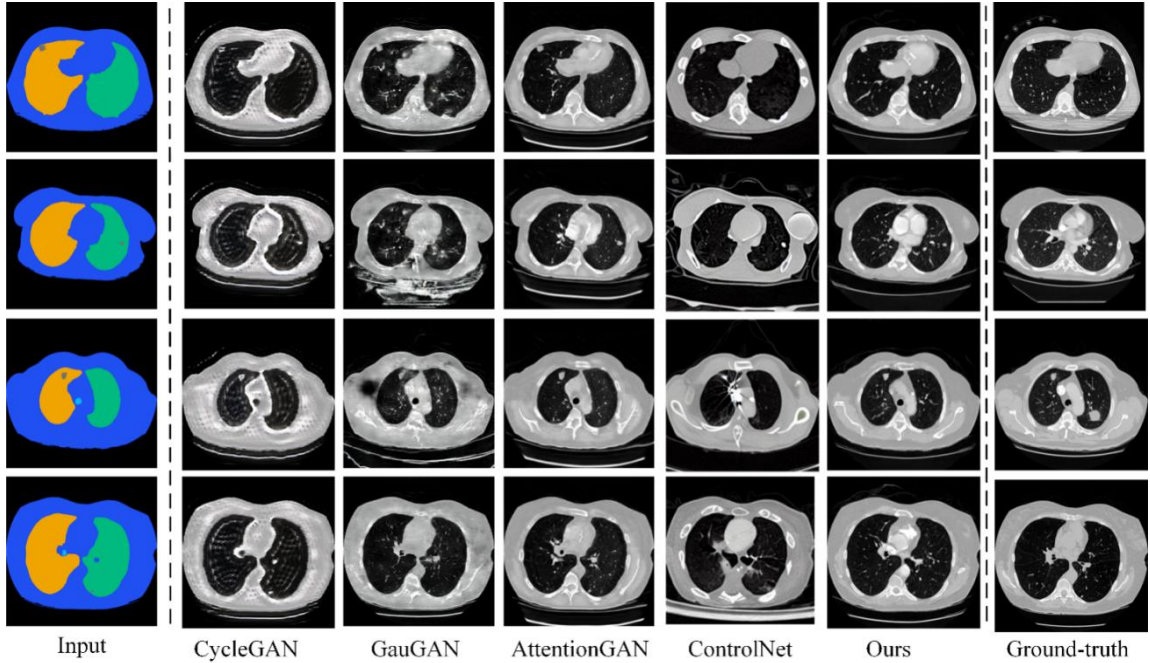

**Fig. 4.** Comparison of generated images under different generation models

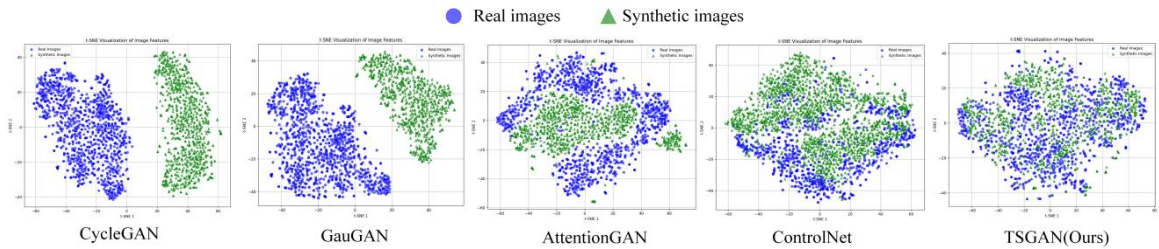

**Fig. 5.** t-SNE dimensionality reduction analysis of the mixture of images generated by different image translation methods and the original images

### 4.4 Experimental results and analysis

**Image Generation Quality Comparison Experiment.** Figure 5 provides a clear comparison of the differences between the images generated by different methods and the original real images. Although CycleGAN and GauGAN can generate lung nodule images with corresponding shapes based on the input mask image, CycleGAN exhibits repetitive textures in the lung parenchyma area and cannot accurately generate the tissue background. GauGAN, on the other hand, has many black artifacts, and the images of the lung area are not clear enough. Furthermore, the lung nodule images exhibit a problem of single texture features; AttentionGAN[26] and ControlNet[27] produce results that are relatively similar to those generated in this paper, but in terms of controlling the size of lung nodules, AttentionGAN occasionally exceeds the mask region, which may affect the controllability of lung nodule generation. ControlNet generates lung nodule regions that are not clearly defined, occasionally losing texture information in this area, and cannot effectively generate details of smaller lung nodules. In contrast, the method proposed in this paper can synthesize lung nodule images that are visually more robust and realistic.

To further evaluate the alignment capabilities of different generation methods, t-SNE visualization was used to compare the feature distributions of synthetic images and original images. As shown in Figure 5, the feature distributions of the synthetic data generated by CycleGAN and GauGAN are independent and separated from those of the real data, indicating that the images generated by these two models lack fidelity. In terms of image presentation, CycleGAN fails to capture the background tissue features of lung nodule images during generation. Although GauGAN's SPADE module enhances spatial control through semantic masks, it lacks sufficient modeling of multi-scale anatomical topology, resulting in significant differences between the generated distribution and real data. In contrast, AttentionGAN achieves local feature alignment by introducing an attention mechanism, reducing the gap between the synthesized data clusters and the real data, but there is still a clear distribution boundary. ControlNet and the method proposed in this paper generate images in the latent space that exhibit a distribution similar to that of real images, but the method proposed in this paper achieves a higher degree of mixing.

As shown in Table 1, CycleGAN and GauGAN perform relatively poorly in terms of FID, PSNR, and SSIM metrics. Although AttentionGAN and ControlNet are similar to our method in terms of generation effects, they are inferior to our method in terms of image generation evaluation metrics.

In summary, the proposed model significantly outperforms CycleGAN, GauGAN, AttentionGAN, and ControlNet in terms of fidelity, controllability, and distribution alignment capability in lung nodule image generation.

**Table 1.** Comparison of image indicators of different generative models

| Model | Full Image | | | Masked region | | |
|---|---|---|---|---|---|---|
| | FID↓ | PSNR↑ | SSIM↑ | FID↓ | PSNR↑ | SSIM↑ |
| CycleGAN[18] | 281.662 | 15.974 | 0.270 | 135.784 | 20.179 | 0.781 |
| GauGAN[19] | 101.574 | 10.306 | 0.336 | 26.130 | 16.200 | 0.361 |
| AttentionGAN[26] | 89.581 | 19.411 | 0.590 | 32.189 | 21.037 | 0.818 |
| ControlNet[27] | **40.086** | 9.960 | 0.305 | 105.391 | 17.155 | 0.687 |
| TSGAN | 58.077 | **20.692** | **0.636** | **25.148** | **21.364** | **0.827** |

**Detection Performance Comparison Experiment.** To compare the impact of different image generation methods on detection performance, three metrics were used to evaluate the generation algorithms, with this experiment conducted on the YOLOv8 model.

As shown in Table 2, the proposed method achieved the best overall performance in terms of recall and mAP. Compared to the original dataset, TSGAN improves YOLOv8's precision, recall, and mAP by 4.6%, 4.1%, and 4%, respectively. Although ControlNet achieves the best metrics in terms of precision, its recall rate decreases compared to before data augmentation. Significantly outperforming other generation algorithms, this demonstrates that the proposed method has a clear advantage in terms of data quality and model detection performance improvement.

**Image Generation Quantity Ablation Experiment.** To validate the effectiveness of the data generated by the method described in this paper, Table 3 shows the performance of two detection models, YOLOv5 and YOLOv8, when the training set is expanded using different amounts of generated data.

The precision, recall, and mAP of YOLOv5 show a clear upward trend as the amount of synthetic data increases, reaching a peak at 2,000 synthetic images. When further increased to 2,500 images, performance slightly decreases, indicating that when the number of synthetic images approaches the number of original images, some noise is introduced into the detection model, leading to a performance decline. YOLOv8 achieves optimal performance when 2,000 synthetic images are added, although precision still slightly increases when 2,500 are added.

In summary, the images generated by the proposed method can effectively improve the performance of the detection model, and the best results are achieved when 2,000 synthetic data points are added in this experiment.

**Table 2.** Effect of image quality generated by different generation algorithm models on detection performance

| Model | Precision | Recall | mAP |
|---|---|---|---|
| No data augmentation | 0.791 | 0.720 | 0.752 |
| CycleGAN[18] | 0.813 | 0.732 | 0.760 |
| GauGAN[19] | 0.744 | 0.788 | 0.785 |
| AttentionGAN[26] | 0.837 | 0.725 | 0.790 |
| ControlNet[27] | **0.845** | 0.7 | 0.789 |
| TSGAN(Ours) | 0.837 | **0.761** | **0.792** |

**Table 3.** Detection performance of using YOLO with different amounts of generated data

| Detector | Data quantity | Precision | Recall | mAP |
|---|---|---|---|---|
| YOLOv5 | real | 0.784±0.013 | 0.725±0.009 | 0.767±0.013 |
| | real +500 | 0.816±0.017 | 0.726±0.022 | 0.772±0.017 |
| | real +1000 | 0.827±0.018 | 0.731±0.010 | 0.780±0.010 |
| | real +1500 | 0.830±0.007 | 0.743±0.020 | 0.772±0.017 |
| | real+2000 | **0.838±0.018** | **0.762±0.024** | **0.790±0.024** |
| | real+2500 | 0.833±0.016 | 0.757±0.015 | 0.785±0.019 |
| YOLOv8 | real | 0.791±0.035 | 0.720±0.007 | 0.752±0.018 |
| | real+500 | 0.831±0.026 | 0.737±0.027 | 0.786±0.026 |
| | real+1000 | 0.833±0.010 | 0.744±0.016 | 0.785±0.017 |
| | real+1500 | 0.836±0.013 | 0.755±0.031 | 0.788±0.015 |
| | real+2000 | 0.837±0.007 | **0.761±0.015** | **0.792±0.019** |
| | real+2500 | **0.839±0.012** | 0.744±0.011 | 0.790±0.010 |

**Ablation Experiments on Different Modules.** To validate the effectiveness of the method proposed in this paper, StyleGAN, the LIA module, and the DWMH module were gradually added to the model for ablation experiments, using YOLOv8 as the baseline network.

As shown in Table 4, using Pix2Pix to generate data improves recall and mAP but does not improve precision. Introducing StyleGAN to generate mask maps enhances image quality and overall detection performance. After adding the LIA module and DWMH module, some metrics reached optimal levels. The final model, which integrates all components, achieves the best overall results in terms of image fidelity and detection performance.

Based on the experimental results, under the same experimental conditions, incorporating attention mechanisms and using a two-stage process to generate lung nodule images both provide some improvement in detection performance.

Table 4. Quantitative results of ablation experiments

| Model | FID↓ | PSNR↑ | SSIM↑ | Precision | Recall | mAP |
|---|---|---|---|---|---|---|
| No data augmentation | - | - | - | 0.791 | 0.720 | 0.752 |
| Pix2Pix | 42.840 | 8.606 | 0.665 | 0.774 | 0.739 | 0.793 |
| Pix2Pix+LIA | 33.217 | 8.468 | 0.660 | 0.806 | 0.733 | 0.797 |
| Pix2Pix+DWMH | 32.958 | 10.514 | 0.660 | 0.804 | 0.738 | 0.8 |
| StyleGan+Pix2Pix | 30.567 | 8.69 | 0.664 | 0.808 | 0.733 | 0.794 |
| StyleGan+Pix2Pix+LIA | **30.264** | 8.678 | 0.668 | 0.806 | 0.749 | **0.798** |
| StyleGan+Pix2Pix+LIA+DWMH | 32.187 | **15.571** | **0.669** | **0.837** | **0.761** | 0.792 |

# Conclusion

This paper addresses the issues of sample imbalance and insufficient diversity in lung nodule CT data by proposing a two-stage generative adversarial network (GAN) framework. By decoupling the lung nodule generation process into two stages—mask generation and image translation—the framework enables controlled synthesis of nodule morphology and texture. The segmentation mask images generated using StyleGAN provide guidance for lung nodule image generation, enhancing image diversity while ensuring controllability. The DL-Pix2Pix model, by incorporating LIA attention and DWMH modules, significantly strengthens the spatial correlation between nodule edge textures and surrounding tissues. Experimental validation shows that the lung nodule images generated by the proposed method outperform traditional generation methods in terms of visual quality and detection task adaptability, improving the lung nodule detection accuracy on the LUNA16 dataset to 83.7% and the mAP metric to 79.2%, confirming the role of synthetic data in enhancing model generalization capabilities.

   Given the limited availability of publicly accessible lung nodule CT datasets, future work will explore using other medical datasets for training and testing to further validate the framework's generalization performance across different anatomical structures and pathological features.